\pdfoutput=1
\documentclass[11pt,a4paper]{article}
\usepackage[hyperref]{acl2020}
\usepackage{times}
\usepackage{latexsym}
\usepackage[ruled,vlined,noend]{algorithm2e}
\usepackage{enumitem}

\usepackage{xspace}
\usepackage{amsfonts}
\usepackage{multirow}
\usepackage{multicol}

\usepackage{microtype}
\usepackage{booktabs}
\usepackage{graphicx}
\usepackage{xcolor}
\usepackage{subfigure}

\aclfinalcopy 

\newcommand\aspect[1]{\mbox{``\textit{#1}''}\xspace}
\newcommand\our{\mbox{ARYA}\xspace}
\newcommand\smallsection[1]{\noindent\textbf{#1.}}

\title{User-Guided Aspect Classification for Domain-Specific Texts}

\author{
   Peiran Li$^1$ $\qquad$ Fang Guo$^3$ $\qquad$ Jingbo Shang$^{1,2}$ \\
  \small $^1$ Department of Computer Science and Engineering, University of California San Diego, CA, USA\\
  \small $^2$ Hal\i c\i o\u glu Data Science Institute,University of California San Diego, CA, USA \\
  \small $^3$ Department of Computer Science, University of Illinois at Urbana-Champaign, IL, USA \\
  \small \texttt{$^{1,2}$\{pel047,jshang\}@ucsd.edu $\quad^3$fangguo1@illinois.edu} \\
}
\date{}

\begin{document}
\maketitle
\begin{abstract}
    Aspect classification, identifying aspects of text segments, facilitates numerous applications, such as sentiment analysis and review summarization.
To alleviate the human effort on annotating massive texts, in this paper, we study the problem of classifying aspects based on only a few user-provided seed words for pre-defined aspects.
The major challenge lies in how to handle the noisy \aspect{misc} aspect, which is designed for texts without any pre-defined aspects.
Even domain experts have difficulties to nominate seed words for the \aspect{misc} aspect, making existing seed-driven text classification methods not applicable. 
We propose a novel framework, \our, which enables mutual enhancements between pre-defined aspects and the \aspect{misc} aspect via iterative classifier training and seed updating.
Specifically, it trains a classifier for pre-defined aspects and then leverages it to induce the supervision for the \aspect{misc} aspect.
The prediction results of the \aspect{misc} aspect are later utilized to filter out noisy seed words for pre-defined aspects.
Experiments in two domains demonstrate the superior performance of our proposed framework, as well as the necessity and importance of properly modeling the \aspect{misc} aspect.

\end{abstract}

\section{Introduction}

Aspect classification is a fundamental task in text understanding, aiming at identifying aspects of text segments~\cite{he2017unsupervised}.
It can facilitate various downstream applications, including sentiment analysis and product review summarization.
For instance, understanding aspects of a product's review sentences can help to deliver a holistic summary of this product without missing any important aspect~\cite{angelidis2018summarizing}.

Following the supervised paradigm to extract aspects requires extensive human effort on annotating massive domain-specific texts, because aspects vary across domains.
For example, in restaurant reviews, possible aspects include \aspect{food}, \aspect{service}, and \aspect{location}.
When it comes to laptop reviews, aspects become \aspect{battery}, \aspect{display}, etc.
Therefore, to alleviate such effort, we study the problem of \textit{user-guided aspect classification}, which only relies on very limited supervision --- only a small number (e.g., 5) of seed words per aspect.

The major challenge of this problem lies in how to handle the \aspect{misc} aspect.
The \aspect{misc} aspect is designed to capture two types of text segments which makes it noisy: 
(1) text segments about some specific aspects out of the pre-defined scope, which are quite common in the real world, 
and (2) text segments talking nothing about any specific aspect (e.g., ``This is one of my favorite restaurants.'').
Due to this noisy nature, even domain experts have difficulties to nominate seed words for the \aspect{misc} aspect, making existing seed-driven text classification methods~\cite{agichtein2000snowball,riloff2003learning,kuipers2006bootstrap,tao2015doc2cube,meng2018weakly} not applicable here.
In this paper, we aim to better incorporate the \aspect{misc} aspect into user-guided aspect extraction.

We make two intuitive, crucial observations, which shed light on the development of our proposed framework.
First, given a text segment, if its distribution over pre-defined aspects is flat, it likely belongs to the \aspect{misc} aspect.
This provides us a chance of inducing \aspect{misc}-aspect supervision from the classifier trained for pre-defined aspects.
Second, given a word, if it is a strong indicator of the \aspect{misc} aspect, it is unlikely to be a good seed word of any pre-defined aspect.
Excluding such words from the seed words of pre-defined aspects would reduce ambiguity, thus becoming a wise decision.

\begin{figure*}
    \centering
    \includegraphics[width=0.95\linewidth]{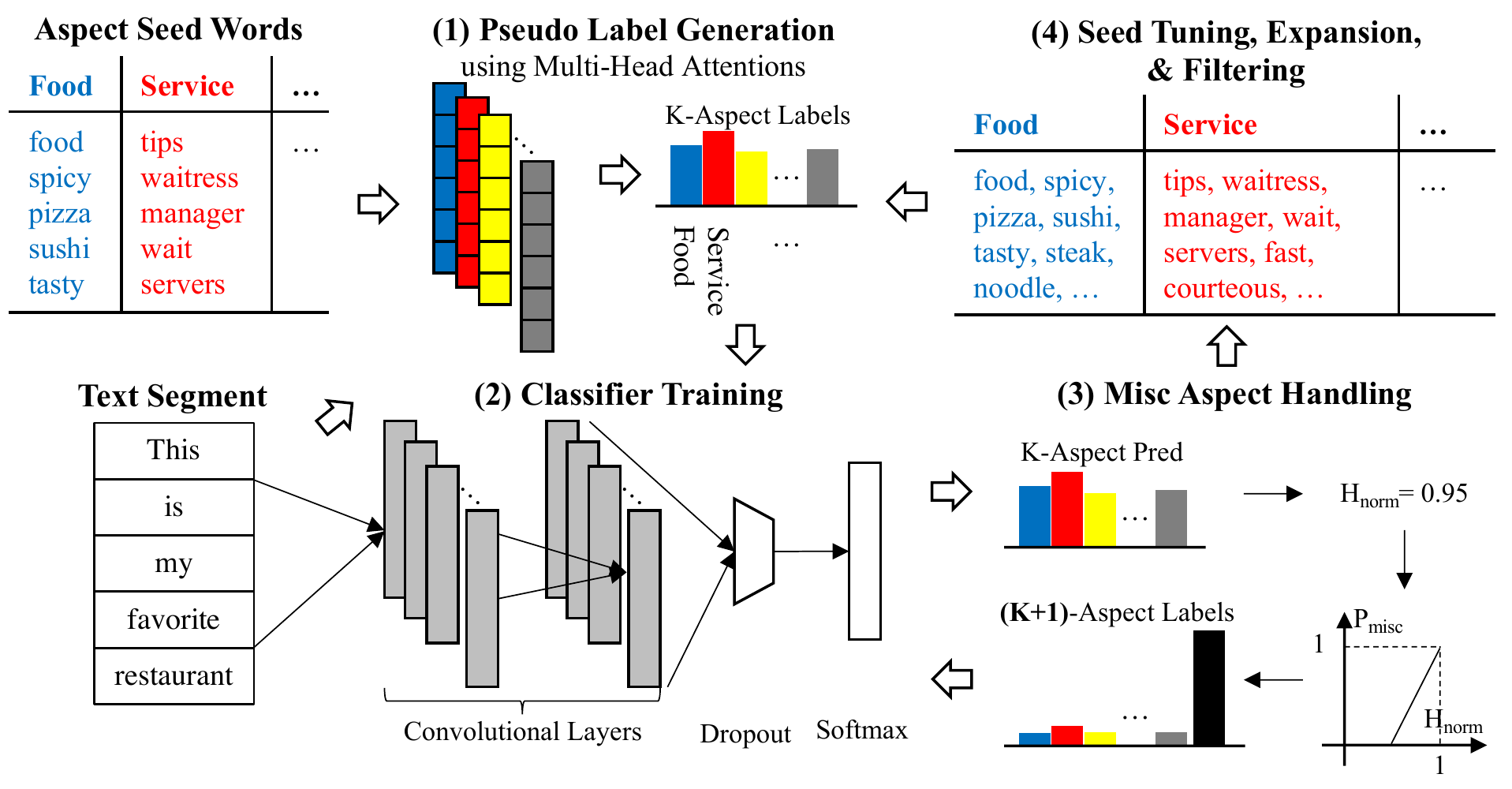}
    \vspace{-3mm}
    \caption{Overview of our proposed framework \our. 
    It enables mutual enhancements between the pre-defined aspects and the \aspect{misc} aspect via iterative classifier training and seed updating.
    Pre-defined aspects help to induce supervision for the \aspect{misc} aspect;
    The \aspect{misc} aspect helps to filter out noisy seed words for pre-defined aspects.
    }
    \label{fig:overview}
    \vspace{-3mm}
\end{figure*}

Acknowledging these observations, we propose a novel framework incorporating the \aspect{misc} aspect in a systematic manner.
As shown in Figure~\ref{fig:overview}, it is an iterative framework, alternatively training the classifier for all aspects and updating seed words of pre-defined aspects.
We name it as \our.\footnote{Our framework is named after \textit{Arya Stark} in Game of Thrones, who kills the Night King, bringing an end to the \textit{Others} (i.e., White Walkers, wights, etc.) forever.}
More specifically, we first train a classifier for K pre-defined aspects based on user-provided seed words.
This K-aspect classifier further induces supervision for the \aspect{misc} aspect based on normalized entropy estimation, enabling a (K+1)-aspect classifier.
This (K+1)-classifier facilitates our comparative analysis, which updates seed words of pre-defined aspects using strong aspect-indicative words.
The predicted \aspect{misc} aspect information is utilized to ensure those noisy words will never appear in seed words of pre-defined aspects.
As one can see here, \our achieves mutual enhancements between pre-defined aspects and the \aspect{misc} aspect.

To our best knowledge, we are the first to systematically handle the \aspect{misc} aspect in user-guided aspect extraction.
Our main contributions are:
\begin{itemize}[leftmargin=*,nosep]
    \item We identify the keystone towards user-guided aspect extraction as the noisy \aspect{misc} aspect.
    \item We develop \our based on two intuitive observations, making pre-defined aspects and the \aspect{misc} aspect mutually enhance each other.
    \item Experiments in two domains demonstrate the superiority of \our and the necessity of considering the \aspect{misc} aspect systematically.
\end{itemize}
\textbf{Reproducibility}. We will release our code and datasets in our GitHub repository\footnote{\url{https://github.com/peiranli/ARYA}}.
\section{Overview}

\smallsection{Problem Formulation}
Given a domain-specific corpus $\mathcal{D}$ of $n$ text segments $\{S_1, S_2, \ldots, S_n\}$, $K$ pre-defined aspects $\{A_1, \ldots, A_K\}$, and a small number of seed words per aspect $\{V_{A_1}, \ldots, V_{A_K}\}$, this paper aims to build an aspect classifier for domain-specific text segments.
A domain here refers to a relatively consistent category of products or services, such as the \textit{hotel} domain, the \textit{restaurant} domain, and the \textit{laptop} domain.

In this paper, we assume that there is at most one specific aspect in each text segments. 
In practice, one can always segment the text in a fine-grained way to ensure that this assumption holds.
In other words, for any input text segment $S_i$, our classifier aims to predict its corresponding aspect label $y_i$.
$y_i$ is either an ID of the pre-defined aspects (between 1 and K) or the number K+1 denoting that $S_i$ focuses on none of the pre-defined aspects.

\begin{algorithm}[t]
\caption{Overall Algorithm}\label{alg:overall}
\SetAlgoLined
  \textbf{Input}: A corpus $\mathbf{D}$ of $n$ text segments $\{S_1, S_2, \ldots, S_n\}$, user-provided seed words for K pre-defined aspects $\{V_{A_1},...,V_{A_K}\}$. \\
  \textbf{Output}: A (K+1)-aspect classifier. \\
  Train word embedding $\mathbf{e}_w$ on $\mathcal{D}$. \\
  \While{\emph{Seed words are not converged}}{
    Compute aspect embedding $\mathbf{a}_j$ (Eq~\ref{eq:aspect_emb}) \\
    Get K-aspect supervision $q_{i,j}$ (Eq~\ref{eq:soft}) \\
    Train K-aspect classifier $M_{K}$ (Sec~\ref{sec:cnn}) \\
    Get (K+1)-aspect supervision $\hat{q}_{i,j}$ (Eq~\ref{eq:soft_k1}) \\
    Tune, expand, and filter seed words (Sec~\ref{sec:seed}) \\
  }
  \textbf{Return} The last (K+1)-aspect classifier.
\end{algorithm}

\smallsection{Our Framework}
\our is an iterative framework as illustrated in Figure~\ref{fig:overview} and Algorithm~\ref{alg:overall}.
In each iteration, we apply the following four steps in order.
\begin{itemize}[leftmargin=*, nosep]
    \item \textbf{Pseudo Label Generation}.
        Given seed words for K aspects, we generate K-aspect pseudo labels for all text segments in the raw corpus.
    \item \textbf{Classifier Training}. 
        We train a K-aspect classifier based on the generated pseudo labels. 
        Our framework is compatible with all text classifiers. 
        As an illustration, we choose to use 1-D CNN in this paper. 
        We will brief its neural architecture for the self-contained purpose.  
    \item \textbf{Misc Aspect Handling}.
        We leverage the predictions of the trained K-aspect classifier to produce pseudo labels for the \aspect{misc} aspect. 
        After that, we train a new (K+1)-aspect classifier, which makes an end-to-end aspect extraction.
    \item \textbf{Seed Tuning, Expansion, and Filtering}. 
        We conduct a comparative analysis to compare and contrast the text segments projected to different aspects to find new and discriminative seed words for each aspect. 
        The \aspect{misc} aspect is utilized here to further filter out noisy seed words for pre-defined aspects.
\end{itemize}

We will discuss the details of the four major components in the following sections.
Before that, here are some basic notations.

\smallsection{Notations}
Each text segment consists of a sequence of tokens, i.e., $S_i = \langle w_1, \ldots, w_{|S_i|} \rangle$, where $|S_i|$ is the number of tokens in $S_i$.
Please note that ``token'' here includes not only single-word words and punctuation but also multi-word phrases (e.g., ``battery life'', ``chocolate cake'') and subword pieces (e.g., ``n't''). 
The tokens are pre-processed from raw texts by applying both tokenization and phrasal segmentation~\cite{shang2018automated}. 

Let $V$ be the vocabulary set of all tokens.
For each token $w \in V$, we denote its $d$-dimensional embedding vector as $\mathbf{e}_w \in {\mathbb{R}}^{d \times 1}$. 
The embedding representation matrix of text segment $S_i$ is then defined as $\mathbf{X}_i = ( \mathbf{e}_{w_1}, \ldots, \mathbf{e}_{w_{|S|}} ) \in \mathbb{R}^{d \times |S_i|}$ by concatenating each row vector.

\section{Pseudo Label Generation}
    We generate pseudo labels following a multi-head attention mechanism, where each attention head focuses on a specific aspect.
    It helps our model focus on aspect indicative words and ignore irrelevant ones, and derive aspect-oriented representation.
    The outputs from all attention heads are finally aggregated to derive the prominent aspect of the text segment.
    
    First, we assume that the user-provided seed words can characterize the aspect's semantics. 
    So we compute $\mathbf{a}_j$, the aspect representation of $A_j$, by averaging embedding of its seed words.
    \begin{equation}\label{eq:aspect_emb}
        \mathbf{a}_j = \frac{\sum_{w \in V_{A_j}} \mathbf{e}_w}{|V_{A_j}|}
    \end{equation}
    
    A higher embedding similarity between a word and an aspect implies that the word is more closely related to the aspect, and it should be paid greater attention to.
    Therefore, given a word $w$, its attention weight is defined as its maximum similarity over K aspects.
    \begin{equation}
        \beta_{w} = \max_{j=1}^{K}\{ \mathbf{a}_j^T \mathbf{e}_{w} \}
    \end{equation}
    Since text segments are usually short, we use the average of its tokens following the attention weights as its aspect-oriented representation $\mathbf{z}_i$.
    \begin{equation}
        \mathbf{z}_i = \frac{\sum_{w \in S_i} \beta_w \cdot \mathbf{e}_w}{\sum_{w \in S_i} \beta_w}
    \end{equation}

    Based on the similarity between text segment representation $\mathbf{z}_i$ and aspect representation $\mathbf{a}_j$, we derive the pseudo label assignments as
    \begin{equation} \label{eq:soft}
        q_{i, j} \propto \exp(\mathbf{a}_j^T \mathbf{z}_i)
    \end{equation}
    We normalize $q_{i, *}$ into a label distribution over all K aspects.

\section{Aspect Classifier Training}\label{sec:cnn}

    Our framework is generally compatible with any text classifiers.
    In this paper, we choose to use a 1D-CNN model because the multi-head attention mechanism in our pseudo label generation can be viewed as applying a few corresponding convolutional filters.
    Specifically, every aspect representation $\mathbf{a}_i$ is equivalent to a convolutional filter of window size one.
    
    As mentioned before, we have $\mathbf{X}_i$ as the embedding representation matrix of text segment $S_i$.
    We feed $\mathbf{X}_i$ to our 1D-CNN model, as illustrated in Figure~\ref{fig:overview}.
    Specifically, we employ various filters of window sizes two, three, and four, corresponding to bi-grams, tri-grams, and four-grams. 
    We apply these filters on the input matrix and then add a dropout layer after convolutional layers to alleviate over-fitting.
    Finally, we use a softmax layer to transform the output to probabilities as $P_{i, j}$, denoting the probability of $S_i$ belonging to aspect $A_j$.
    The pseudo label distribution $q_{i}$ generated in the previous step serves as supervision here, using the KL-divergence loss as below. 
    \begin{equation}
        \mathcal{L} = KL(q_i, P_i) = - \sum_{j = 1}^{K} q_{i, j} \log \frac{P_{i, j}}{q_{i, j}}
    \end{equation}

    The same classification logic applies to the training of both K-aspect and (K+1)-aspect classifiers.

\begin{figure}[t]
    \centering
    \subfigure[Restaurant Dataset]{
        \includegraphics[width=0.45\linewidth]{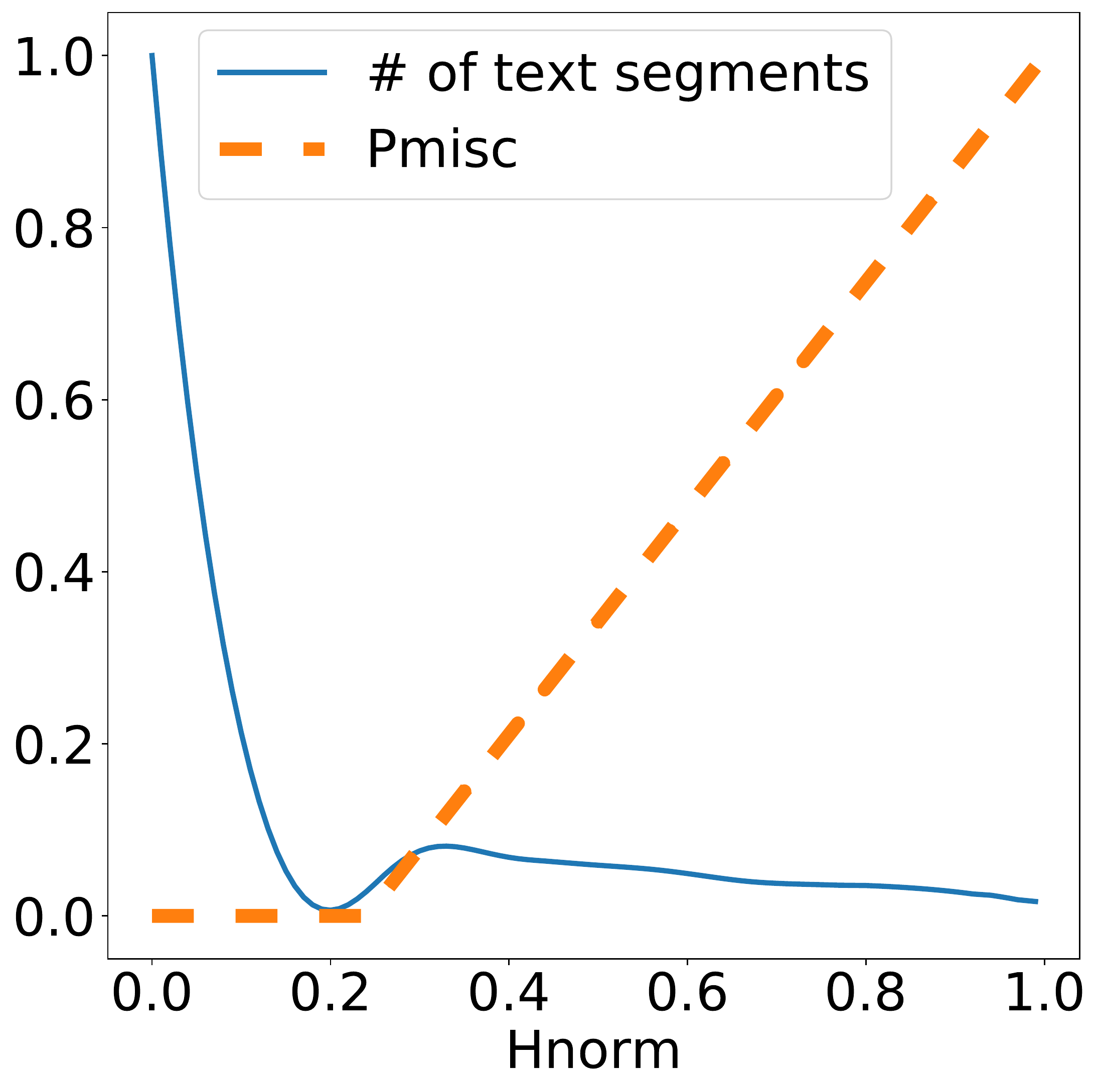}
    }
    \subfigure[Laptop Dataset]{
        \includegraphics[width=0.45\linewidth]{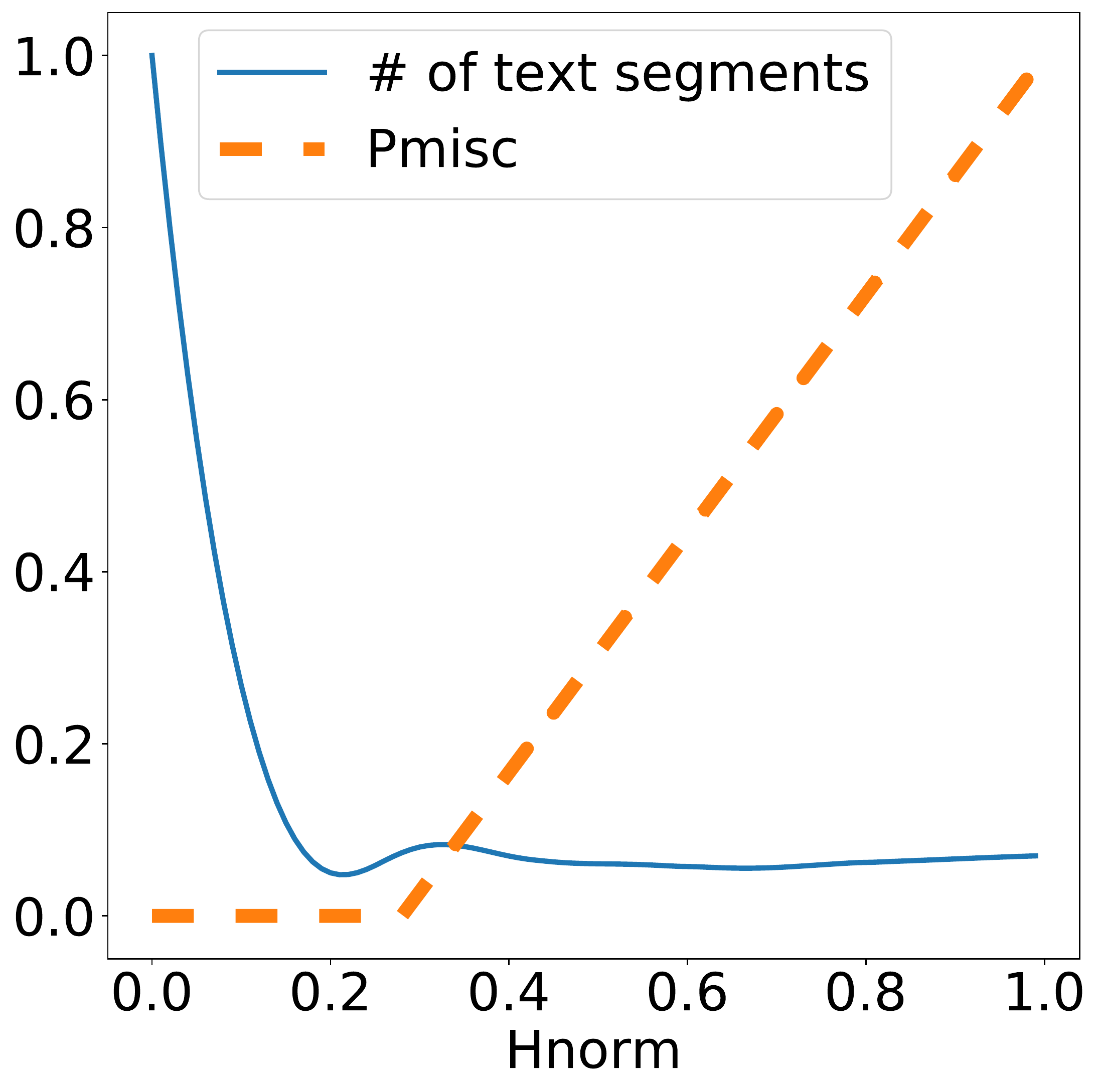}
    }
    \vspace{-3mm}
    \caption{$H_{norm}$ Distribution and $P_{misc}$ Visualization.}
    \label{fig:h_norm}
    \vspace{-3mm}
\end{figure}

\section{Misc Aspect Handling}
    In aspect extraction, two types of text segments belong to the \aspect{misc} aspect:
    (1) text segments about some specific aspects different from the K pre-defined aspects;
    and (2) text segments talking nothing about any specific aspects.
    These text segments are expected to have a relatively flat distribution in the predictions of the K-aspect classifier.
    Therefore, it is intuitive to leverage normalized entropy $H_{norm}$, which measures how chaotic the distribution is, to estimate the likelihood of $S_i$ belonging to the \aspect{misc} aspect, i.e., $P_{misc}$. 
    Specifically,
    \begin{equation}
        H_{norm} = -\frac{1}{\log K} \sum_{j = 1}^{K} P_{i, j} \log(P_{i, j})
    \end{equation}
    As shown in Figure~\ref{fig:h_norm}, we plot the distribution of $H_{norm}$ for all text segments on both the Restaurant and Laptop datasets. 
    One can easily observe that a large volume of the text segments have low $H_{norm}$, indicating that they belong to some pre-defined aspects.
    At the same time, those \aspect{misc} aspect text segments follow a long-tail distribution over large $H_{norm}$ values.
    Ideally, we want to (1) classify text segments with low enough $H_{norm}$ values to be a non-\aspect{misc} and (2) assign a higher $P_{misc}$ if the $H_{norm}$ is higher.
    Therefore, we propose to leverage a ReLU-like function to quantify $P_{misc}$ based on $H_{norm}$.
    \[
        P_{i, misc} = \left\{ 
        \begin{array}{cc}
            (H_{norm} - \gamma) / (1-\gamma) &  H_{norm} \ge \gamma\\
            0 & H_{norm} < \gamma \\
        \end{array} \right.    
    \]
    We choose the value of $\gamma$ as the 3rd quantile of the $H_{norm}$ scores of all documents because based on Figure~\ref{fig:h_norm}, 3rd quantile will give a suitable pivot point.
    Specifically, in Figure~\ref{fig:h_norm}, the $\gamma$ values on the Restaurant and Laptop datasets are 0.24 and 0.28, respectively. 

    
    After getting this, we combine $P_{i, misc}$ and $P_{i, j}$:
    \begin{equation}\label{eq:soft_k1}
    \hat{q}_{i, j} = \left\{ 
        \begin{array}{cl}
            (1 - P_{i, misc}) P_{i, j} & j \le K\\
            P_{i, misc} & j = K+1 \\
        \end{array} \right.    
    \end{equation}
    
    
    Finally, we obtain the pseudo labels $\hat{q}_{i, j}$ for all aspects, including the misc aspect.
    We then train a (K+1)-aspect 1D-CNN classifier.
    
\section{Seed Tuning, Expansion, and Filtering}\label{sec:seed}
    Besides the user-provided seed words, there are usually more strong aspect indicator words embedded in the raw input corpus.
    It could be helpful to discover and add such words into the seed sets.
    
    \smallsection{Seed Tuning}
    Not every word could be a candidate seed word (e.g., stopwords).
    Therefore, we build a candidate pool based on the K-aspect classifier.
    Specifically, we try to replace each word by the special UNK token and compute the KL divergence between the prediction results before and after. 
    Given a word, if there exists one text segment where this word leads to a KL divergence difference more than 0.05, the word becomes a candidate. 
    The intuition here is we want to prepare a candidate pool with high recall and reasonable precision.
    Also, as further ranking and filtering will be applied, this threshold is fairly easy to decide.


    \smallsection{Seed Expansion}
    Then, we expand the seed sets by ranking and adding words from the candidate pool.
    Given an aspect $A_j$ and its candidate pool $C_j$, we mainly consider two measurements:
    \begin{itemize}[leftmargin=*,nosep]
        \item \textbf{Indicative}. 
        As the pseudo label generation process can be viewed as a soft version of string matching using embedding, we want to select words whose presence strongly indicate a certain aspect. 
        Mathematically, we want to select the word $w$, if it has a high posterior probability $P(A_j | w)$.
        $P(A_j | w)$ means that given the presence of a word $w$, how likely the text segment belongs to the aspect $A_j$.
        Therefore, we define the indicative measure as 
            \begin{equation}
                \mbox{Indicative}(A_j, w) = \frac{f_{A_j, w}}{f_{A_j}}
            \end{equation}
        where $f_{A_j, w}$ is the frequency of the word $w$ appeared in text segments of the aspect $A_j$, and $f_{A_j}$ refers to the total text segments of the aspect $A_j$.
        The frequency is calculated based on the prediction results on the training set.
        \item \textbf{Distinctive}.
        Ideally, a seed word should be only frequent in its own aspect.
        Therefore, we propose a distinctive measure to capture this.
        It measures how distinctive this word $w$ in aspect $A_j$ is compared with all other aspects.
            \begin{equation}
                \mbox{Distinctive}(A_j, w) = \frac{f_{A_j, w}}{\max_{k \neq j} f_{w, A_k}}
            \end{equation}
    \end{itemize}
    Since these two scores are of different scales, we aggregate them using the geometric mean, which has been shown effective in other comparative analyses~\cite{tao2015doc2cube}. 
    Ranking by the aggregated score, we replace the seed words of the aspect $A_j$ by the top words here.
    
    \smallsection{Seed Filtering}
    It is worth noting that the same ranking heuristic can be applied to the \aspect{misc} aspect as well.
    We observe that highly ranked words in the \aspect{misc} aspect are mostly general words or some noisy words that are related to multiple pre-defined aspects.
    By checking some examples on the Restaurant dataset,
    we observe that ``restaurant'' is ranked high in the \aspect{misc} aspect, as it can appear in text segments of any aspects;
    the word ``place'' is also a top-ranked word for the \aspect{misc} aspect.
    Other than location-related text segments, it also appears frequently in text segments like ``this restaurant is such a great place.''
    Intuitively, the user may select this word as a seed word for \aspect{location} aspect, however, it is in fact very noisy.
    Therefore, when replacing the seed words, we propose to maintain a new pool of noisy words following the ranking in the \aspect{misc} aspect and exclude top words in this pool from seed words in pre-defined aspects.
\begin{table}[t]
    \centering
    \caption{Dataset Statistics.}\label{tbl:dataset}
    \vspace{-0.3cm}
    \small
    \begin{tabular}{ccc}
        \toprule
        \textbf{Dataset} & \textbf{Unlabeled Segments} & \textbf{Test Segments} \\
        \midrule
        Restaurant & 16,061 & 1,166 \\ 
        \midrule
        Laptop & 14,683 & 780 \\
        \bottomrule
    \end{tabular}
    \vspace{-0.3cm}
\end{table}

\section{Experiments}

In this section, we empirically evaluate our proposed framework \our against many compared methods.
We also explore the effects of the number of iterations and the number of seeds.
A case study about seed word evolution will be presented too.

\subsection{Datasets}

    We have prepared two review datasets in the restaurant and laptop domains for evaluation. 
    Table~\ref{tbl:dataset} presents you some statistics.
    These two datasets can be found in our repo\footnote{\url{https://github.com/peiranli/ARYA}}.
    \begin{itemize}[leftmargin=*,nosep]
        \item \textbf{Restaurant.} 
            There are 5 aspects in our Restaurant dataset: \aspect{food}, \aspect{service}, \aspect{ambience}, \aspect{drinks}, and \aspect{location}.
            For training, we have collected 16,061 unlabeled restaurant reviews from the Yelp Dataset Challenge data\footnote{\url{https://www.yelp.com/dataset/challenge}}.
        \item \textbf{Laptop.}
            There are 7 aspects in our Laptop dataset: \aspect{support}, \aspect{display}, \aspect{battery}, \aspect{software}, \aspect{keyboard}, \aspect{os}, \aspect{mouse}. 
            For training, we are using 14,683 unlabeled Amazon reviews on laptop, collected by~\citet{mcauley2015image,he2016ups}.
    \end{itemize}

\begin{table}[t]
    \centering
    \caption{User-Provided Seed Words for the Restaurant Dataset. By default, we randomly sample 5 seed words from each aspect and run experiments.}
    \label{tab:Seed words List1}
    \vspace{-3mm}
    \small
    \begin{tabular}{l l}
        \toprule
        {\bf Aspect }&{\bf Seed Word List}\\
        \midrule
        \multirow{2}{*}{\textit{Location}} & street, convenient, block, avenue, river, \\
        & subway, neighborhood, downtown, bus  \\
        \midrule
        \multirow{2}{*}{\textit{Drinks}} & drinks, beverage, wines, margaritas, sake,\\
        & beer, wine list, cocktail, vodka, soft drinks\\
        \midrule
        \multirow{2}{*}{\textit{Food}} & food, spicy, sushi, pizza, tasty, \\ 
        &steak, delicious, bbq, seafood, noodle\\
        \midrule
        \multirow{2}{*}{\textit{Ambience}} & romantic, atmosphere, room, seating, small, \\
        & spacious, dark, cozy, quaint, music\\
        \midrule
        \multirow{2}{*}{\textit{Service}} & tips, manager, wait, waitress, servers, \\
        & fast, prompt, friendly, courteous, attentive\\ 
        \bottomrule
    \end{tabular}
    \vspace{-0.3cm}
\end{table}

\begin{table*}[t]
    \centering
    \caption{Evaluation Results on the Restaurant and Laptop Datasets. All precision, recall, and F$_1$ scores are averaged in the macro-weighted manner. Underlines highlight the best compared models.}\label{tbl:F1}
    \vspace{-2mm}
    \small
    \begin{tabular}{cccccccc}
        \toprule
         & \multicolumn{3}{c}{\textbf{Restaurant}} & \multicolumn{3}{c}{\textbf{Laptop}} \\ 
        \cmidrule{2-7}
        \textbf{Method} & Precision & Recall & F$_1$ & Precision & Recall & F$_1$ \\
        \midrule
        CosSim & 0.5455 & 0.4782 & 0.4985 & 0.6055 & 0.5437 & 0.5083\\ 
        ABAE & 0.5494 & 0.4904 & 0.5112 & 0.6127 & 0.6168 & 0.5950 \\ 
        MATE & 0.5613 & 0.5127 & 0.5177 & 0.6418 & 0.6550& 0.6474 \\ 
        WeSTClass & 0.6153 & 0.5259 & \underline{0.5461} & 0.6688 &	0.6848 & \underline{0.6523}  \\
        Dataless & 0.5225 & 0.4467 & 0.4265 &0.5601  &0.5693  & 0.5569\\ 
        BERT & 0.5955 & 0.5285 & 0.5404 & 0.5949 & 0.5672 & 0.5632 \\
        \midrule
        Best+OurMisc & 0.5864 & 0.5373& 0.5256 & 0.6724 & 0.6996 & 0.6685 \\
        \our & \textbf{0.7410} & \textbf{0.6913} & \textbf{0.7067} & \textbf{0.7849} & \textbf{0.7321} & \textbf{0.7447}\\
        \midrule
        \our-NoIter & 0.6934 & 0.6740 & 0.6749 & 0.7508 & 0.7037 & 0.7027\\ 
        \our-NoTuning & 0.7019 & 0.6620 & 0.6729 & 0.7349 & 0.6874 & 0.6822\\ 
        \our-NoFilter & 0.7145 & 0.6706 & 0.6836 & 0.7619 & 0.7158 & 0.7306\\ 
        \bottomrule
    \end{tabular}
    \vspace{-0.3cm}
\end{table*}
    
    \smallsection{User-Provided Seed Words}
    For both datasets, we ask three domain experts to provide 10 seed words for each pre-defined aspect. 
    Table~\ref{tab:Seed words List1} shows the seed word list provided by one expert for the Restaurant dataset.
    By default, we will randomly choose 5 seed words from them to train all the models, including both ours and baselines.
    We report the average of these test results.
    For one tricky aspect, the \aspect{keyboard} aspect of the Laptop dataset, we have only collected 3 seed words.
    
    \smallsection{Pre-processing}
    We pre-process the corpus using the spaCy\footnote{\url{https://spacy.io/}}.
    Special characters such as "*", "\#" and redundant punctuations are removed.
    We learn word embedding on the unlabeled training corpus.

\subsection{Compared Models}
    We compare our model with a wide range of baseline models, described as follows.
    \begin{itemize}[leftmargin=*,nosep]
        \item \textbf{CosSim} assigns the most similar aspect to each text segment according to the cosine similarity between the average word embedding of the text segment and the average word embedding of all seeds in each aspect.
        \item \textbf{Dataless}~\cite{song2014dataless} accepts aspect names as supervision and leverages Wikipedia and Explicit Semantic Analysis (ESA) to derive vector representation of both aspects and documents. 
        The class is assigned based on the vector similarity between aspects and documents.
        \item \textbf{ABAE}~\cite{he2017unsupervised} is an unsupervised neural topic model. 
        We extend the ABAE by utilizing user-provided seed words for each aspect to align its topics to pre-defined aspects.
        \item \textbf{MATE}~\cite{angelidis2018summarizing} is an extended version of ABAE, which accepts seed information for guidance and replaces ABAE's aspect dictionary with seed matrices.
        \item \textbf{WeSTClass}~\cite{meng2018weakly} is the state-of-the-art weakly supervised text classification model, which accepts seed words as supervision. 
        \item \textbf{BERT}~\cite{devlin2018bert} is a powerful contextualized representation learning technique. 
        We use seed words matching and majority voting to generate sentence labels and then fine-tune the BERT for classification.
    \end{itemize}
    Most of these models do not take care of the \aspect{misc} aspect systematically. 
    Therefore, we fine-tune the best compared method using our proposed \aspect{misc}-aspect handling, referred as \textbf{Best+OurMisc}.
    
    We denote our model as \textbf{\our}. 
    In addition, we have a few ablated versions as follows.
    \textbf{\our-NoIter} uses our proposed \aspect{misc} aspect handling technique to generate the probability of \aspect{misc} aspect based on K-aspect classifier, however, without any further steps. 
    \textbf{\our-NoTuning} refers to the version of our model without the seed tuning technique, i.e., no KL divergence threshold for seed word candidates.
    \textbf{\our-NoFilter} is our model without the seed filtering technique, i.e., no noisy seed words removal in pre-defined aspects based on \aspect{misc} aspect information.

\subsection{Experiment Setup}
    
    \smallsection{Default Parameters}
        We set the word embedding dimension $d = 200$. 
        For the classifier training, we fix the number of epoch as $5$ since the training error tends to converge after 5 epochs.
        The KL divergence threshold for seed tuning is set to $0.05$. This value is set based on some human efforts. One can easily observe that words lead to a KL divergence difference less than $0.05$ are not very representative for that aspect.
        Based on the raw corpus sizes, we set the maximum number of seed words per each aspect as $10$ on the Restaurant dataset, and $8$ on the Laptop dataset. 

    \smallsection{Evaluation Metrics}
        We use macro-weighted average precision, recall, and F1 scores.

\begin{table*}[t]
\caption{Seed Word Evolution Examples. The 0-th iteration indicates the user-provided seeds.}\label{tbl:seed_evolve}
\vspace{-3mm}
\center
\small
\begin{tabular}{cccl}
    \toprule
    \textbf{Dataset} & \textbf{Aspect} & \textbf{Iter} & \multicolumn{1}{c}{\textbf{Seed Words}} \\
    \midrule
    \multirow{6}{*}{\textbf{Restaurant}} & \multirow{2}{*}{food} & 0 & spicy, pizza, sushi, food, tasty  \\
    & & 1 & pizza, spicy, variety, tasty, tuna, sushi, portion, food, specials, bland \\
    \cmidrule{2-4}
    & \multirow{4}{*}{location} & 0 & avenue, convenient, river, street, block  \\
    & & 1 & located, block, view, convenient, river, avenue \\
    & & 2 & located, block, street, view, convenient, park, river, avenue \\
    & & 3 & located, block, street, view, convenient, park, river, york, avenue \\
    \midrule
    \multirow{5}{*}{\textbf{Laptop}} & \multirow{3}{*}{keyboard} & 0 & keyboard, key, space  \\
    & & 1 & keyboard, keys, key\\
    & & 2 & keys, keyboard, numeric, volume, palm, key, layout, keyboards \\
    \cmidrule{2-4}
    & \multirow{2}{*}{os}& 0 & system, os, ios, windows, mac  \\
    & & 1 & system, os, ios, operating, mac, windows, lion, interface\\
    \bottomrule
\end{tabular}
\end{table*}

\subsection{Experiment Results}
    We present the evaluation results on the Restaurant and Laptop datasets in Table~\ref{tbl:F1}.
    It is clear that our proposed method \our outperforms all other methods with significant margins on both datasets because none of these models considers the \aspect{misc} aspect systematically.
    Even compared with the fine-tuned second best models Best+OurMisc, \our results in 18\% and 8\% in absolute improvements over it on the Restaurant and Laptop datasets, respectively.
    It is also worth noting that \our-NoIter significantly outperforms all compared methods.
    All these observations show the importance of properly handling the \aspect{misc} aspect.
    
    Among all compared methods, MATE is arguably the second-best method.
    It utilizes the multi-head attention mechanism, which is the same as our pseudo label generation step.
    This implies that attention mechanism is very important for aspect extraction tasks. 
    \our generalizes attentions to more convolutional filters, thus being able to train a more powerful model.

    The advantage of \our over \our-NoIter demonstrates the importance of progressively refine the model by updating seed words at every iteration.
    Comparing \our-NoTuning and \our-NoIter, one can see that if we do not carefully limit the scope of seed word candidates, there is a risk of adding noisy seed words that will lead to even worse performance (e.g., on the Laptop dataset).
    The improvement of \our over \our-NoFilter reveals the effectiveness of filtering the seed words in pre-defined aspects by the \aspect{misc} aspect.


\begin{figure}[t]
    \centering
    \includegraphics[width=0.49\linewidth]{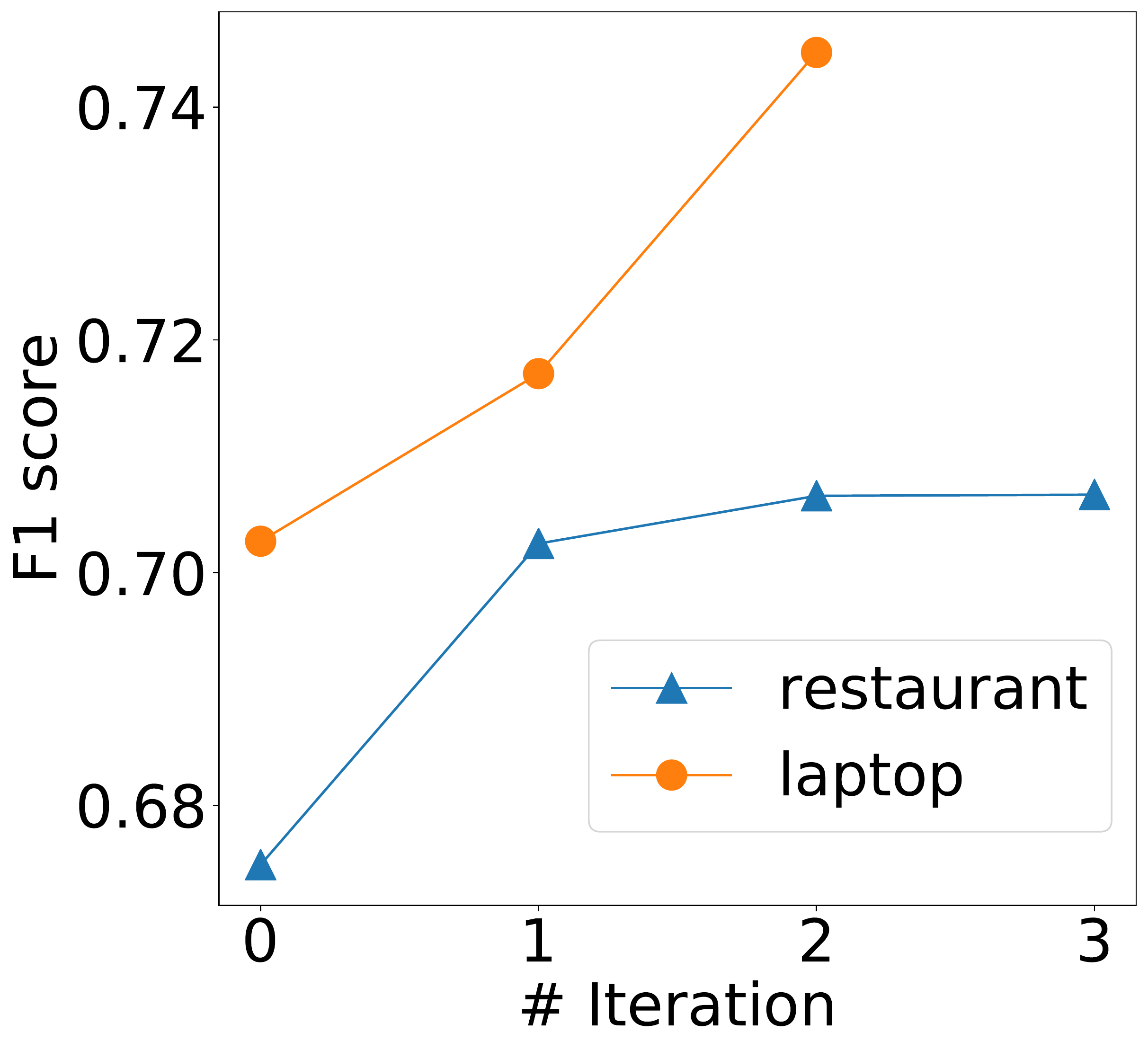}
    \vspace{-3mm}
    \caption{F$_1$ scores in Different Iterations. \our keeps iterating until the seed words converge.}
    \label{fig:iterations}
\end{figure}

\subsection{Seed Word Evolution}

\our keeps iterating until the seed words converge.
So the number of iterations in \our is decided automatically.
Figure~\ref{fig:iterations} shows that the F$_1$ score increases w.r.t. iterations on both datasets.
This suggests that our framework truly enables mutual enhancements between pre-defined aspects and the \aspect{misc} aspect over iterations.

Table~\ref{tbl:seed_evolve} presents the seed words of each aspect w.r.t. different iterations on both datasets.
We can observe that the seed words become much better after the seed expansion than the initial seed words. 

As mentioned before, even domain experts feel challenging to provide seed words for the \aspect{keyboard} aspect. 
Only three seed words, ``keyboard'', ``key'', and ``space'' are given at the very beginning.
After a few rounds of seed tuning, expansion, and filtering,
some interesting words are added to its seed set, such as ``layout'', ``numeric'', and ``palm'', which make sense for the keyboard aspect.
For example, ``palm'' describes the how comfortable the palms are when typing on a keyboard or how big the keyboard is compared with palms.
It is interesting to see that our model can automatically discover these words beyond typical examples come up by experts.

We also observe that the seed word sets of popular aspects converge faster than infrequent aspects.
For example, on the Restaurant dataset, the \aspect{food}, \aspect{ambience}, and \aspect{service} aspects converge after the 1st iteration, and the \aspect{drinks} and \aspect{location} aspects requires 2 and 3 iterations, respectively.
The first three have significantly more text segments than the latter two.

Another observation is that the tricky aspects converge slower than the other aspects. 
For example, on the Laptop dataset, the \aspect{keyboard} aspect converges much slower than the other aspects, because it is very counter-intuitive to come up with the seed words such as ``palm'' and ``numeric''. 
On the contrary, the \aspect{os} aspect is relatively easy compared with other aspects. 



\subsection{Misc Text Segment Examples}

    We present two successfully classified text segments of the different types of \aspect{misc} aspect.
    
    The first example is from the Restaurant dataset, ``There is nothing more pleasant than that.'', without any specific aspect.
    \our detects that the word ``pleasant'' as a noisy word, because it can refer to \aspect{service} or \aspect{ambience}. 
    Therefore, it is filtered for these two aspects.
    Eventually, \our predicts the probabilities of this segment belong to \aspect{misc}, \aspect{service}, and \aspect{ambience} as 0.38, 0.29, and 0.25 respectively.
    Therefore \aspect{misc} wins in the end.
    
    The second example is from the Laptop dataset: ``the only problem is that i had to add 1 gb RAM, the computer was kinda slow.'', about the out-of-pre-defined \aspect{hardware} aspect.
    \our predicts it as \aspect{misc} and \aspect{os} with chances 0.47 and 0.19 respectively, mainly because the word ``slow'' is widely used to complain about OS.
\section{Related Works}

Aspect extraction was originated at a document-level task, instead of working on text segments.
Rule-based methods~\cite{hu2004mining,liu2005opinion,zhuang2006movie,scaffidi2007red,zhang2010extracting,qiu2011opinion} are the pioneers along this direction.
A number of unsupervised learning methods based on the LDA topic model and its variants~\cite{titov2008modeling,zhao2010jointly,brody2010unsupervised,mukherjee2012aspect,zhang2016hybrid,shams2017enriched} treat extracted topics as aspects.
More recently, a neural model ExtRA~\cite{luo2018extra} is proposed to further improve the aspect extraction at the document level.
However, since our problem focuses on text segments, directly applying these document-level methods leads to some unsatisfactory results.

There are several recent unsupervised attempts on aspect extraction for text segments.
ABAE~\cite{he2017unsupervised} employs an attention module to learn embedding for text segments and an auto-encoder framework to build aspect dictionaries.
However, it requires users to first set the number of topics as a much larger number than the number of desired aspects, and then manually merge and map the extracted topics back to the aspects. 
Building upon ABAE, \citet{angelidis2018summarizing} further proposed a multi-seed aspect extractor MATE using seed aspect words as guidance.
This model keeps the human effort at a minimal degree and fits our problem setting well.
However, even with its multi-task counterpart, the reconstruction objective in MATE model is not able to provide adequate training signals.
Our proposed method leverages the seed word tuning and expansion to overcome this issue, thus outperforming MATE significantly in the extensive experiments.

Our problem shares certain similarities with the weakly-supervised text classification problem.
Existing methods can build document classifiers by taking either hundreds of labeled training documents~\cite{tang2015pte,miyato2016adversarial,xu2017variational}, class/category names~\cite{song2014dataless,li2018unsupervised}, or user-provided seed words~\cite{meng2018weakly} as the source of weak supervision.
However, all these methods assume that users can always provide seeds for all classes, while overlooking the noisy \aspect{misc} aspect in our problem.
We incorporate the \aspect{misc} aspect systematically into our framework.
\section{Conclusions and Future Work}

In this paper, we explore to build an aspect extraction model for text segments using only a few user-provided seed words per aspect. 
We identify the key challenge lies in how to properly handle the \aspect{misc} aspect, for which even domain experts cannot easily design seed words. 
We propose a novel framework, \our, which incorporates the \aspect{misc} aspect systematically.
In our framework, we induce supervision for the \aspect{misc} aspect using seed words of pre-defined aspects.
At the same time, we utilize the \aspect{misc} aspect information to filter out the noisy words from the seed list of pre-defined aspects.
Extensive experiments have demonstrated the effectiveness of \our and verified the necessity of modeling the \aspect{misc} aspect. 

In the future, we would like to integrate the extracted aspect information with downstream tasks, such as sentiment analysis and opinion summarization.
We also want to explore the use of contextualized representation in weakly supervised aspect extraction, further disambiguating words based on contexts. 
In addition, we are interested in extending our work to document classifications even with multiple labels per document.


\bibliography{acl2020}
\bibliographystyle{acl_natbib}

\end{document}